\documentclass[sigconf, nonacm]{acmart}
\usepackage{tabularx}
\usepackage{graphicx}
\usepackage{multirow} 
\usepackage{subcaption}
\usepackage{xcolor}
\usepackage{booktabs}
\usepackage{caption}
\usepackage[most]{tcolorbox}
\AtBeginDocument{%
  }

\begin{document}
\title{IPGeoAI: Transformer-Based Geolocation with LLM Semantic Fusion}

\author{Avinash Kadimisetty}
\email{avinashk@meta.com}
\affiliation{%
  \institution{Meta}
  \city{Menlo Park}
  \state{CA}
  \country{USA}
}
\author{Andy Jinqing Yu}
\email{andyyu@meta.com}
\affiliation{%
  \institution{Meta}
  \city{Menlo Park}
  \state{CA}
  \country{USA}
}

\author{Philip Favaloro}
\email{pfavaloro@meta.com}
\affiliation{%
  \institution{Meta}
  \city{Boston}
  \state{MA}
  \country{USA}
}
\author{Wenlong Liu}
\email{wenlonl@meta.com}
\affiliation{%
  \institution{Meta}
  \city{Boston}
  \state{MA}
  \country{USA}
}
\author{Xiaolu Xiong}
\email{beardeer@meta.com}
\affiliation{%
  \institution{Meta}
  \city{Boston}
  \state{MA}
  \country{USA}
}

\begin{abstract}
Accurate city-level IP Geolocation is an important enabler for the modern digital ecosystem, underpinning services ranging from local content delivery and targeting to digital rights enforcement. However, traditional heuristic and database-driven methods often struggle to resolve the complex, non-linear allocation patterns of modern network infrastructures, particularly within the exploding IPv6 address space and transient mobile networks. In this paper, we introduce IPGeoAI, a novel deep learning model architecture that reframes geolocation from a static lookup problem to a sequential modeling task. Our approach utilizes the Transformer Encoder to capture hierarchical dependencies inherent in IP subnet structures. We propose a method to resolve geographic ambiguity by integrating unstructured semantic context via a Zero-Shot LLM Feature Extraction pipeline. We utilize Large Language Models to transform raw, noisy Autonomous Systems (AS) descriptions into structured, domain-specific metadata (such as 'University' vs. 'ISP' or 'Global' vs. 'Local') via an offline pre-computation process. By fusing these semantic signals into the network via a Multi-Head Cross-Attention module, we bridge the gap between numerical network topology and real-world semantic identity. Extensive offline evaluation on a proprietary dataset spanning 200,000 cities demonstrates that IPGeoAI significantly outperforms a leading external vendor in city-level granularity. By adopting a hierarchical inference strategy that refines coarse-grained country signals, our model achieves a 6\% improvement in city-level accuracy while extending coverage to 100\% of the traffic.  Furthermore, in large-scale online production tests, the model drove a statistically significant +0.35\% improvement in our 1st-tier downstream use cases metric. We conclude by discussing  model serving flow and presenting ablation studies that isolate the specific gains attributed to semantic fusion and attention mechanisms.
\end{abstract}




\maketitle

\section{Introduction}
\label{sec:introduction}
An IP address is a unique identifier assigned to each internet-connected device, acting as both a virtual location anchor and a key element of user identity. IP Geolocation --- the process of mapping an IP address to a physical location --- remains one of the most persistent and intricate challenges in internet measurement. It is an important capability of the modern digital ecosystem: content delivery networks (CDN) rely on it for latency optimization; digital rights management (DRM) systems use it to enforce regional licensing; and security platforms leverage it as a primary source for fraud detection. In the context of a user, it is the ubiquitous signal when GPS is unavailable or restricted. While resolving an IP address to coarse-grain Country or Region/State level has arguably become a solved problem with high reliability, achieving city-level precision represents a significant frontier in internet geolocation intelligence. Furthermore, network or internet service providers do not allocate IP addresses to specific cities. Instead, they designate blocks of IP address ranges that span multiple regions encompassing numerous cities. This practice introduces a significant challenge in accurately mapping an IP address to a city, as this mapping can fluctuate over time, necessitating continuous data updates to maintain predictive accuracy. 

Given the ubiquity of IP signals, they serve as a primary location enabler for three core verticals:

\begin{itemize}
    \item \textbf{Personalized experiences}: IP geolocation maintains content relevance when precise GPS data is unavailable. 

    \item \textbf{Event correlation for measurement/analytics}: It provides the spatial context needed to correlate disparate events and ensure measurement accuracy.

    \item \textbf{Local Discovery and Marketplaces}: Platforms rely on accurate default views to minimize user friction. Incorrect localization (e.g., showing San Francisco listings to a San Jose user) significantly degrades relevance and the user experience.
\end{itemize}

Given the scale, complexity and importance of this problem, learning intricate patterns of both IPv4 and IPv6 addresses is important. The difficulty of this task stems from the inherent orthogonality between internet topology and physical geography. Early research focused on active measurement techniques such as constraint-based geolocation \cite{gueye2006constraint} and topology-based methods like Octant \cite{wong2007octant} or Spotter \cite{laki2011spotter}. These systems rely on sending probes (pings, traceroutes) from distributed landmarks to triangulate a target's position based on latency constraints. While theoretically sound, active measurement faces scalability barriers in the modern internet: it is intrusive, computationally expensive, and increasingly thwarted by firewalls and middleboxes that drop ICMP traffic. Furthermore, the relationship between network delay and geographic distance is often non-linear due to circuitous routing policies, limiting the precision of these methods to the regional level rather than the city level \cite{poese2011ip}. Commercial providers rely on aggregating various data sources—including WHOIS registries and user-submitted data—to build static lookup tables. While effective for stable, high-traffic IPv4 networks (the "head" of the distribution), these heuristic databases suffer from severe limitations in coverage (recall) for the "long tail." They fundamentally rely on memorization: if a prefix has not been seen with sufficient frequency, it cannot be located. This leaves a significant blind spot for new allocations, transient mobile networks, and the burgeoning IPv6 landscape, where the vastness of the address space renders exhaustive surveying impossible. Recent attempts to apply machine learning to this domain have largely treated it as a simple classification problem using static features. For example, Eriksson et al. \cite{eriksson2010learning} and Shavitt et al. \cite{shavitt2010structural} proposed neural networks trained on delay measurements or basic WHOIS attributes. However, these models typically treat IP addresses as static identifiers, ignoring the hierarchical sequential nature of network addressing, and lack the capacity to resolve the semantic ambiguity of complex organizations.

IP geolocation in large scale platforms is commonly driven by heuristic aggregation pipelines. These systems process raw location signals on a daily cadence, synthesizing historical data to establish mappings for the widest possible range of IP addresses. While effective for stable, high-traffic IPv4 networks, these heuristic approaches suffer from low coverage - particularly for new IPv6 allocations or transient mobile networks where historical data is sparse. While these methods demonstrate high efficacy on previously encountered IP addresses, they exhibit limited generalization when applied to unseen network addresses. Though the issue is addressed with IP prefix level location mapping, it comes at a loss of precision. Even then, such methods do not have 100\% coverage and continuously rely on commercial data providers for IPs that cannot be mapped. To overcome these limitations, we argue that IP geolocation must evolve from a problem of static database lookup to one of predictive modeling by extracting the sequential patterns present inherently within the IPs. We hypothesize that there are latent, non-linear patterns encoded within the structure of IP addresses themselves - subtle correlations between subnet bits and allocation policies that heuristics fail to capture. In this paper, we propose IPGeoAI, a novel deep learning architecture that treats the IP address as a hierarchical sequence, processed by a Transformer Encoder to capture the nested dependencies in an IP address. Furthermore, recognizing that numerical signals alone are often ambiguous, we introduce a method to integrate unstructured semantic context. To ensure low-latency inference, we leverage LLMs in an offline pipeline to extract rich metadata about Autonomous Systems, distinguishing, for instance, a global cloud provider from a hyper-local municipal ISP, we fuse sequential pattern recognition with semantic reasoning. We demonstrate that this multi-modal approach not only improves upon the precision of legacy heuristics and commercial data providers (referred to as third party baseline hereon) on known traffic but achieves significantly superior generalization on unseen network addresses offering a scalable solution for the IPv6 era.

\section{Related Work}
\label{sec:related_work}
The evolution of Internet geolocation has been driven by the increasing need for location-aware services, fraud detection, and regulatory compliance. The literature can be broadly categorized into static registry-based approaches, active measurement-based techniques, and modern data mining and machine learning frameworks.

\subsection{Registry and Database-Driven Approaches}
\label{sec:related_work_database}
Early geolocation efforts relied heavily on administrative data. WHOIS databases, which store registration information for IP blocks and Autonomous Systems (AS), served as the primary source for initial geolocation attempts \cite{padmanabhan2001investigation}. While fundamental, standard WHOIS lookups are often inaccurate because the registered address of an organization frequently differs from the physical location of its network infrastructure \cite{lu2021whois}. Recent studies have highlighted the transition from WHOIS to the Registration Data Access Protocol (RDAP) to improve data accessibility, though accuracy issues persist \cite{corneo2024whois}. Efforts to attribute historical IP data have also been proposed to support longitudinal studies \cite{streibelt2023whois}.

To address these limitations, commercial vendors developed proprietary databases. These services aggregate data from partner feeds, user submissions, and mining techniques. While widely used, they often suffer from "edge opacity" and lack granularity at the street level \cite{casado2007peering, shavitt2011geolocation}.

\subsection{Measurement and Topology-Based Approaches}
To overcome the granularity limitations of static databases, researchers introduced active measurement techniques that infer location from network delay and topology.

\paragraph{Delay and Constraint-Based Methods}
A seminal work in this domain is Constraint-Based Geolocation (CBG) by Gueye et al. \cite{gueye2004cbg}, which treats delay measurements from probes to a target IP as distance constraints, using multilateration to solve for the location. Recognizing that Internet routing does not always follow straight lines, Katz-Bassett et al. proposed Topology-Based Geolocation (TBG) \cite{katz2006tbg}, which leverages router topology constraints alongside delay measurements to improve estimation. Laki et al. introduced \textit{Spotter} \cite{laki2011spotter}, a probabilistic approach that models the delay-distance relationship to handle network noise more effectively.

\paragraph{Landmark and PoP-Based Methods}
Improving precision often requires "landmarks"—hosts with known locations. Wang et al. \cite{wang2011street} proposed a three-tier framework using web servers as landmarks to achieve street-level accuracy. Others have focused on the Point of Presence (PoP) level; for instance, Yuan et al. \cite{yuan2019pop} developed algorithms to partition network nodes into PoPs using traceroute data. Similarly, Liu et al. \cite{liu2016ip} and Zu et al. \cite{zu2018city} utilized PoP topology analysis to refine city-level geolocation. Innovative approaches have also mined unconventional landmarks, such as public webcams \cite{li2021geocam} or router hostnames \cite{luckie2021learning}, to densify the reference points available for triangulation.

\paragraph{Machine Learning and Network Measurements} As datasets grew, machine 
learning methods emerged to model the complex relationship between 
network measurements and geography. Early approaches, such as, Eriksson et al. \cite{eriksson2010learning}, employed Naive Bayes classifiers trained on latency and hop counts.  Jiang et al. \cite{jiang2016neural} demonstrated that neural networks using stable landmarks could  outperform geometric multilateration, while Hong et al. \cite{hong2023cheap} used machine learning to fill in missing looking-glass data for delay-based methods.
More recently, Graph Neural Networks (GNNs) have achieved state-of-the-art 
results by modeling network topology explicitly. Wang et al. \cite{wang2022connecting} proposed \textit{GraphGeo}, which uses RTT measurements and traceroute-derived graphs to 
model IP host relationships for street-level accuracy. 
Extensions to this paradigm include \textit{GNN-Geo} by Ding et al. \cite{ding2023gnn} and \textit{GraphNEI} by Ma et al. \cite{ma2023graphnei}. To handle the uncertainty inherent in network measurements, Tai et al. introduced \textit{TrustGeo} \cite{tai2023trustgeo} and later \textit{RipGeo} \cite{tai2023ripgeo}, incorporating self-supervised 
learning and perturbation strategies.
While these methods achieve impressive accuracy, they require distributed 
measurement infrastructure (e.g., RIPE Atlas probes) for each prediction, 
introducing latency overhead prohibitive for real-time inference at 
hyper-scale.

\subsection{Paradigm Limitations and Our Innovation} All measurement-based 
approaches, from classical multilateration to modern GNNs, share a 
fundamental constraint: they require active probing infrastructure 
for each prediction, introducing latency overhead and scalability 
limitations. Our work 
bridges this gap by applying deep learning to passively collected 
IP-GPS telemetry. We treat IP addresses as hierarchical sequences 
processed by a Transformer Encoder, and enrich predictions with 
LLM-derived semantic features—enabling high-precision, real-time 
geolocation at hyperscale without measurement infrastructure.

\section{IPGeoAI Methodology}
\label{sec:ipgeoai_methodology}
Our approach departs from the active measurement paradigms discussed in earlier sections, which are operationally intractable at our scale due to the immensity of the IPv6 128-bit address space and the latency overhead of continuous probing. Constrained by the requirements for high-throughput, large scale low latency inference, we instead adopt a passive, data-driven methodology. We re-frame the geolocation not as a static lookup or triangulation task, but as a predictive modeling problem. This allows us to learn latent structural dependencies from historical telemetry and generalize to unseen subnets without the computational cost of active network interaction. We propose a Semantic-Aware Attention Fusion Network, a novel hybrid architecture that reframes IP Geolocation as a sequence modeling problem enhanced by multi-modal semantic fusion. A core contribution of our work is the application of the Transformer architecture \cite{Vaswani2017Attention} to the domain of network addressing. The subsequent sections detail the methodology employed for data collection and aggregation in preparation for model training and evaluation.

\subsection{Data}

\subsubsection{Geo Hierarchy Data:}
\label{sec:geo_hierarchy_data}
Location entity datasets related to boundaries can help determine geo-hierarchy from each location granularity. For example, New York City is part of New York State which is in the United States of America. This geo hierarchy is later used to uplevel locations from city to region and country. All the different cities are later encoded to give a numerical ID hereafter referred to as CityID.

\subsubsection{Ground Truth:} 
\begin{figure}[h]
    \centering
    \includegraphics[height=5cm, keepaspectratio]{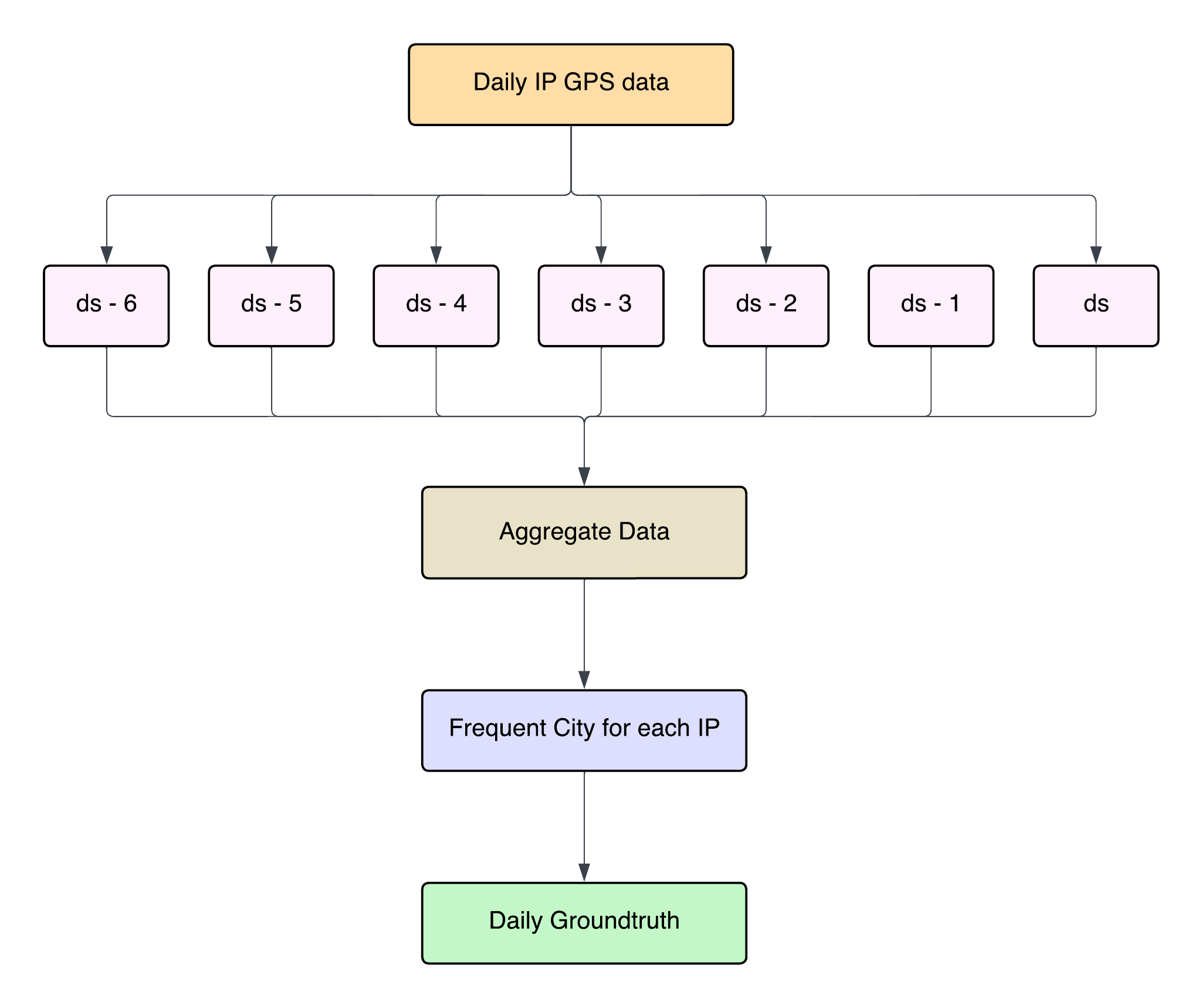}
    \caption{GPS Data Aggregation for Ground Truth (ds denotes today)}
    \label{fig:groundtruth}
\end{figure}

High-frequency IP and GPS logs contain too much fluctuation to be used directly for training models. Pre-processing is therefore required to convert this raw information into reliable ground truth. An important aspect of the proposed methodology for establishing a reliable ground truth for IP Geo-location involves a time-based aggregation strategy. Specifically, to overcome the inherent daily volatility and sparse coverage often found in raw IP GPS data, the system aggregates this daily data over a trailing 7-day period as shown in figure \ref{fig:groundtruth} (7-day sliding window to smooth out transient noise). Rather than relying on the GPS coordinates associated with an IP address on any single day, which might be subject to anomalies, device movement, or temporary network routing changes, the system synthesizes a more stable, representative, and accurate location for that IP. By pooling the data points collected across a full week (the past 7 consecutive days), the system effectively constructs a single, stabilized day of ground truth. This aggregation acts as a powerful filter, dampening transient errors and highlighting the most consistent and frequent physical location associated with the IP address over the recent past, thereby improving the overall accuracy and reliability of the resulting IP Geo stabilized ground truth dataset.

%%%%%%%%%%%%%%%%%%%%%%%%%%%%%%%%%%%%%%%%%%%%%%%%%%%%%%%%%%%%%%%%%%%%%%%%%%%%%%%%%%%%%%%%%%%%%%%%%%%%%%%%%%%%%%%%%%%%%%%%%%%%%

\subsection{Features}
\label{sec:features}
A few features that are available at IP level are the Network Provider Information which is referred to as ASN (Autonomous System Number) hereon and the Connection Type for an IP Address. Information regarding IP-to-ASN mappings and IP connection types is considered publicly available, as these details can be either directly queried from global BGP routing tables or derived from open regional internet registry records \cite{route_views}.  

ASN is a globally unique identifier assigned to an Autonomous System (AS). An AS is a collection of interconnected IP routing prefixes under the control of one or more entities (usually an Internet Service Provider or a large organization) that presents a common, clearly defined routing policy to the Internet. In the context of IP geolocation, the ASN associated with an IP address provides crucial information about the network provider or owner of that IP block. The Connection Type, as its name implies, designates the nature of the network allocation for a given IP address by the network provider. For the purposes of this work, we have consolidated the connection types into three distinct categories: Mobile, Non-Mobile, and Mixed. 

\begin{figure}[h]
    \centering
    \includegraphics[width=0.8\linewidth]{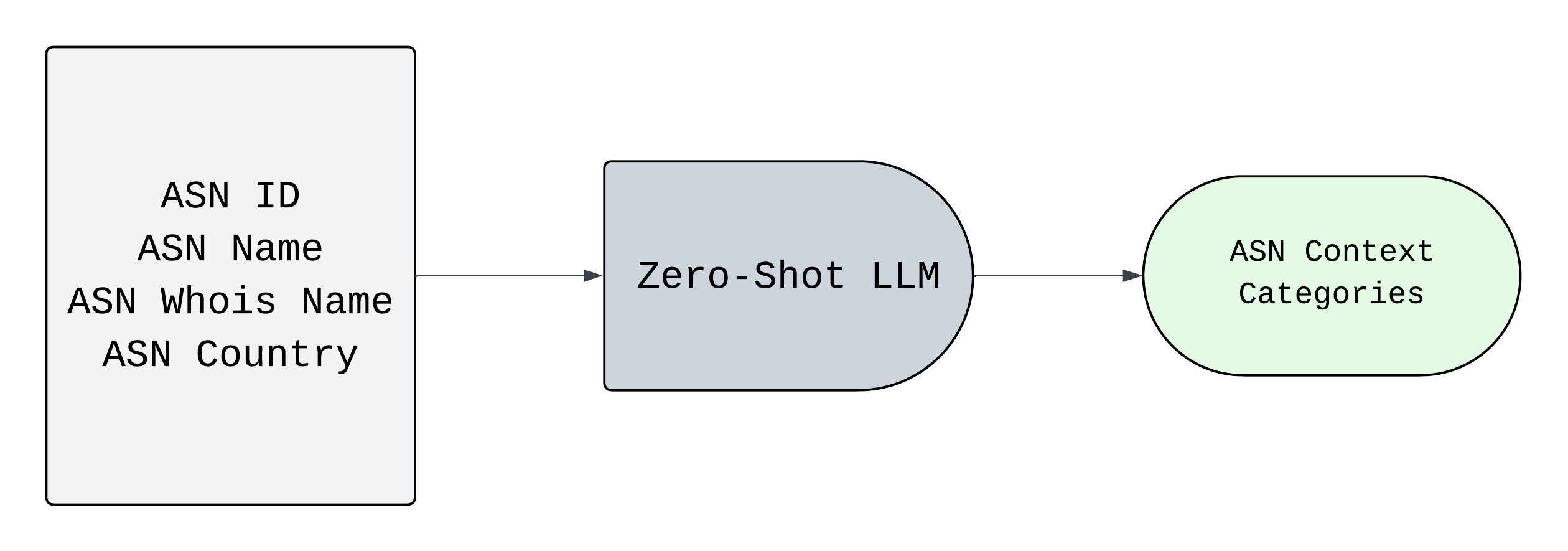}
    \caption{Extracting semantic features from ASN using LLMs}
    \label{fig:llm_features}
\end{figure}

\begin{table*}[t]
\centering
\caption{LLM-derived ASN Feature Descriptions. These features capture semantic world knowledge to aid geolocation inference.}
\label{tab:asn_features}
\begin{tabularx}{\textwidth}{@{}p{3cm} p{3.5cm} X@{}}
\toprule
\textbf{Category} & \textbf{Feature Name} & \textbf{Description \& Rationale} \\
\midrule
% Category 1
\multirow{5}{=}{\textbf{Organizational Identity \& Scope}} 
 & ASN Type & Classifies operational role (e.g., ISP, University) to learn distinct spatial priors (e.g., static campuses vs. distributed mobile pools). \\ \addlinespace
 & ASN Sector & Identifies business sector (e.g., Gov, Edu), providing cues on stability and administrative structure. \\ \addlinespace
 & ASN Scope & Estimates geographical footprint (Local, Global) to regularize predictions for local networks. \\ \addlinespace
 & ASN Multinational & Flags cross-border operations to alert the model to potential country-level ambiguities. \\
\midrule
% Category 2
\multirow{3}{=}{\textbf{Usage \& Demographics}} 
 & ASN Usage Profile & Characterizes primary function (e.g., Residential, Data Center) to distinguish geographic certainty. \\ \addlinespace
 & ASN Urban Focus & Infers population density (Urban, Rural) to resolve conflicts between metros and suburbs. \\
\midrule
% Category 3
\multirow{4}{=}{\textbf{Topology \& Hierarchy}} 
 & ASN Tier Level & Estimates peering hierarchy (Tier 1 vs. Tier 3) to differentiate continental backbones from local stubs. \\ \addlinespace
 & Hosting/Mobile Flags & Explicit flags for infrastructure (Cloud/Hosting) where IP geolocation is often decoupled from physical location. \\
\bottomrule
\end{tabularx}
\end{table*}

\subsubsection{Semantic Enrichment of Network Metadata via LLMs:}

While the ASN and Connection Type features provide critical contextual information, they are high dimensional (the data has around 100,000 ASNs) making them difficult for a standard classification model to interpret effectively. To bridge this gap, we introduce a novel feature extraction pipeline that leverages the reasoning capabilities of LLMs to extract structured metadata from the ASN information as shown in figure \ref{fig:llm_features}.

For each unique ASN in our dataset, we have its unique identifier referred to as ASN ID, the name of the provider referred to as ASN Name, WHOIS name of the ASN and the primary country of the ASN. We then employed a LLM as a zero shot classifier to use the above attributes of ASN to populate a nine dimensional feature vector. This feature vector generated using the expert knowledge of LLMs provides the model with the network's likely topological behaviour. The nine features are categorized into three distinct functional groups - Organizational Identity \& Scope, Usage \& Demographics and Network Topology \& Hierarchy. The nine features are described in Table \ref{tab:asn_features}. To ensure the reliability of the zero-shot LLM predictions, we conducted a manual audit on a random sample of 25 ASNs. We cross-referenced the generated metadata (specifically ASN Usage Profile and ASN Scope) against authoritative public registries (PeeringDB). The LLM demonstrated a 76\% agreement rate with public registries. This high fidelity confirms that the LLM effectively acts as a high-quality "soft" labeler, injecting valid domain knowledge rather than hallucinations.

%%%%%%%%%%%%%%%%%%%%%%%%%%%%%%%%%%%%%%%%%%%%%%%%%%%%%%%%%%%%%%%%%%%%%%%%%%%%%%%%%%%%%%%%%%%%%%%%%%%%%%%%%%%%%%%%%%%%%%%%%%%%%
\subsection{Model Architecture}
\label{sec:model_architecture}
We define the IP City Geolocation task as a multi-class classification problem. Given an input tuple $x = (IP, M)$ where $IP$ represents the network address and $M$ represents a set of associated features (e.g. ASN, Connection Type, LLM based Features etc.), the goal is to predict the City ID $y \in \mathcal{C}$ from a set of $|\mathcal{C}| \approx 200,000$ global cities as the output of the model. Unlike traditional methods that treat the IP address as a static identifier, we model it as a hierarchical sequence. Let $O_{ip} = [o_1, o_2, \dots, o_n]$ be the sequence of octets representing the IP address. This sequential representation allows the model to capture the nested subnet structures inherent to CIDR allocation policies (e.g., recognizing that 192.168.1.0/24 is a subset of 192.168.0.0/16). The metadata $M$ is transformed into a set of dense features via learned embeddings $E$. 

This sequential IP representation along with the learned embeddings $E$ are integrated via a Multi-Head Attention Fusion module. This module dynamically weights auxiliary metadata—including novel semantic features derived from LLMs, allowing the system to resolve ambiguities by learning complex, non-linear dependencies between an IP's numerical structure and its real-world identity. This architecture enables high-precision classification of individual IPs across a global output space of over 200,000 cities. The following sections detail the components of this proposed model design.

\subsubsection{Sequential IP Encoding:}

\begin{figure}[h]
    \centering
    \includegraphics[width=0.9\linewidth]{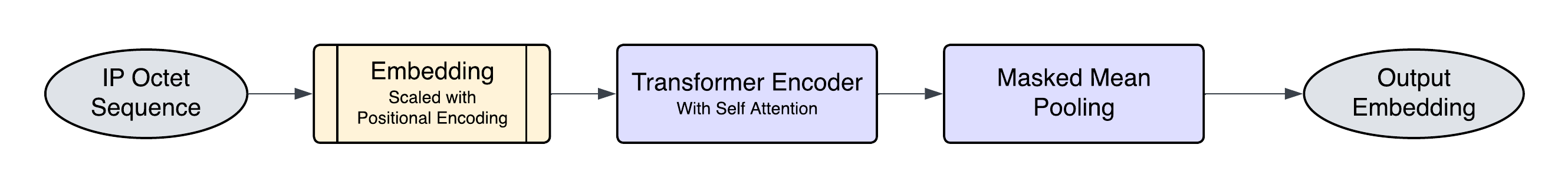}
    \caption{Sequential IP Encoder}
    \label{fig:sequential_ip_encoding}
\end{figure}

Let $\mathcal{I}$ denote the space of valid IP addresses. We define a tokenization mapping: $\phi : \mathcal{I} \to \mathbb{Z}^L$ that transforms an IP address into a sequence of $L$ integer octets $\mathcal{O} = [o_1, o_2, \dots, o_n]$, where $o_t \in [0, 255]$. The sequence length is fixed at $n = 4$ for IPv4 and $n=16$ for IPv6. To capture the sequential properties of the address space, each octet $o_t$ is projected into a $d$-dimensional latent space ($d = 256$) via a learnable embedding matrix $\mathbf{E} \in \mathbb{R}^{256 \times d}$, enriched with a positional encoding $P_t$ to preserve the hierarchical order of the address space: $$\mathbf{h}_t^{(0)} = E(o_t) + P_t$$ This sequence is processed by a Transformer Encoder $\mathcal{T}$ with $N=16$ layers as shown in figure \ref{fig:sequential_ip_encoding}. The self-attention mechanism within $\mathcal{T}$ models the dependencies between octets, learning how the interpretation of lower-order bits (e.g., $o_4$) is conditioned by higher-order network prefixes (e.g., $o_1$). The final IP representation $\mathbf{z}_{ip} \in \mathbb{R}^d$ is obtained by aggregating the output sequence: $$\mathbf{z}_{ip} = \text{MeanPool}(\mathcal{T}(\mathbf{h}_1^{(0)}, \dots, \mathbf{h}_L^{(0)}))$$ This vector $\mathbf{z}_{ip}$ encapsulates the latent structural properties of the IP address, serving as the primary signal for geolocation.

\subsubsection{Semantic Context Encoding:}

\begin{figure}[h]
    \centering
    \includegraphics[width=0.9\linewidth]{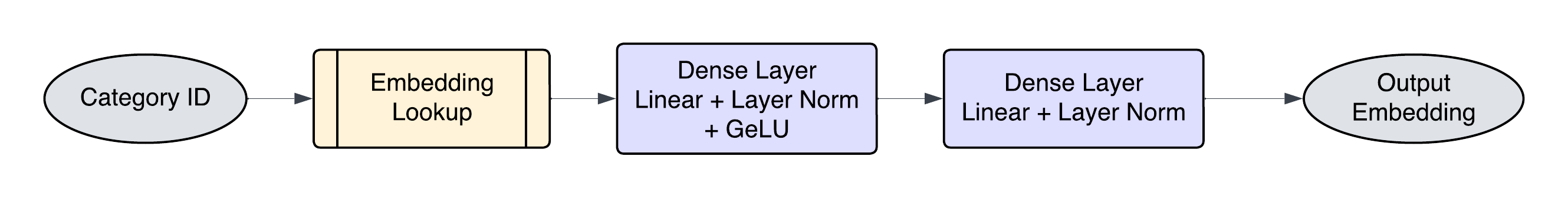}
    \caption{Semantic Context Encoding}
    \label{fig:semantic_context_encoder}
\end{figure}

To integrate unstructured network knowledge, we define a set of categorical metadata features $\mathcal{M} = \{m_{asn}, m_{conn}, m_{type}, m_{sector}, \dots \}$. This set includes the raw ASN ID $m_{asn}$, the connection type $m_{conn}$, and the nine novel semantic attributes derived from LLMs (e.g., asn\_scope, asn\_hosting\_flag). Each feature $m_k \in \mathcal{M}$ is a discrete variable with a vocabulary size $V_k$. We map each feature to a dense vector in the shared latent space $\mathbb{R}^d$ via a dedicated learnable embedding function $E_k: \{1, \dots, V_k\} \rightarrow \mathbb{R}^d$ as shown in figure \ref{fig:semantic_context_encoder}.
$$\mathbf{c}_k = E_k(m_k)$$
The resulting set of context vectors $\mathbf{C} = \{\mathbf{c}_{asn}, \mathbf{c}_{conn}, \mathbf{c}_{type}, \dots \}$ represents the "context bank." By projecting these disparate semantic concepts (e.g., "Global Scope" or "Mobile Infrastructure") into the same high-dimensional space as the IP vector $\mathbf{z}_{ip}$, we enable the subsequent fusion module to compute direct semantic similarities between the network's address structure and its metadata identity.

\subsubsection{Cross-Modal Attention Fusion:}

\begin{figure}[h]
    \centering
    % height=5cm sets the height
    % keepaspectratio ensures the image doesn't look stretched
    \includegraphics[height=8cm, keepaspectratio]{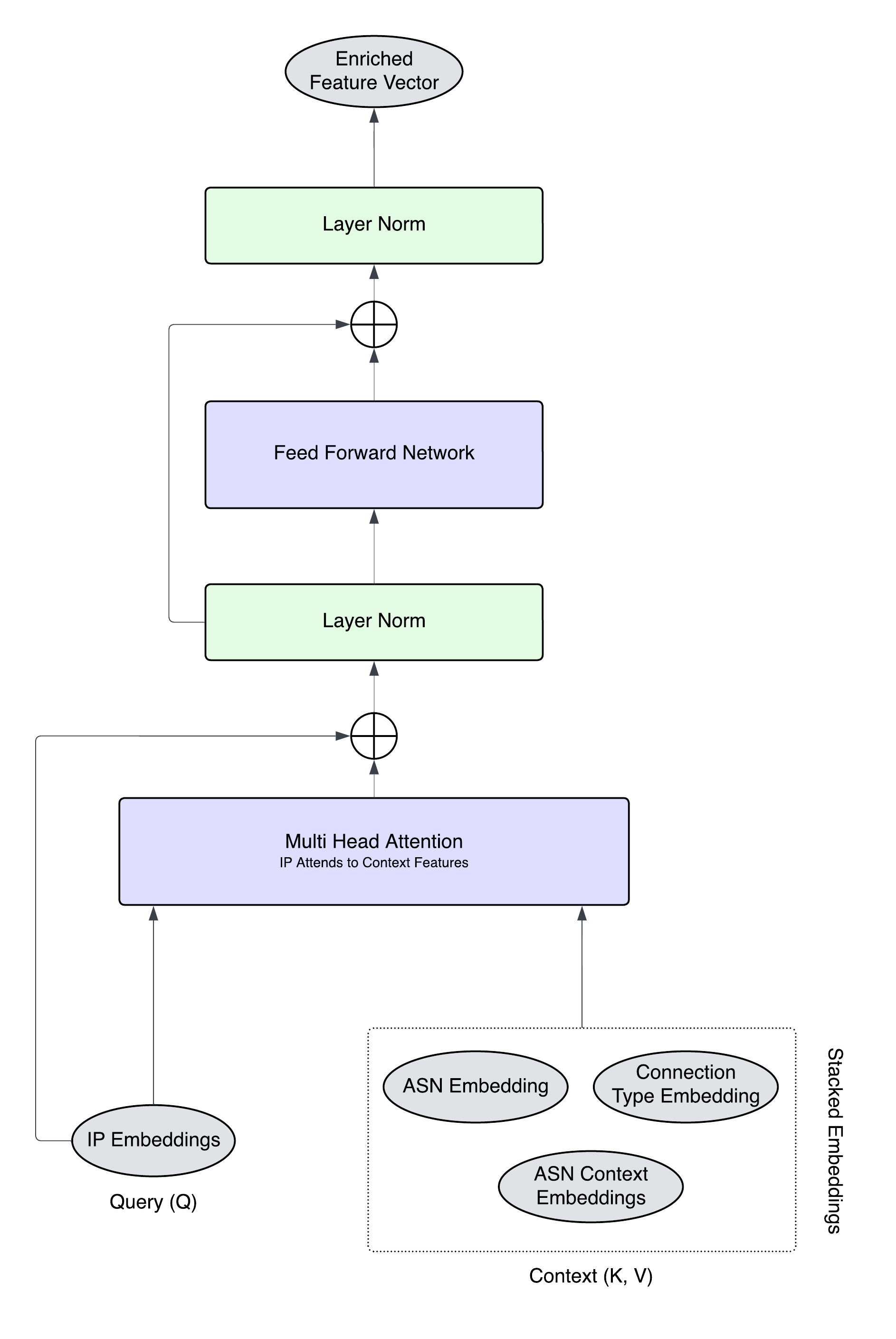}
    \caption{Cross-Modal Attention Fusion}
    \label{fig:attention_fusion} 
\end{figure}

To effectively integrate the numerical IP signal with this diverse semantic context, we propose a Multi-Head Cross-Attention Fusion module as shown in figure \ref{fig:attention_fusion}. Standard feature concatenation imposes a static weight on all inputs, which is suboptimal for geolocation where feature relevance is highly conditional (e.g., asn\_hosting\_flag is important for cloud IPs but irrelevant for residential ones).

Our module treats the encoded IP vector $\mathbf{z}_{ip}$ as the Query ($Q$), and the set of semantic context embeddings $\mathbf{C}$ as the Keys ($K$) and Values ($V$). This formulation allows the IP signal to dynamically attend to specific metadata features based on its own structural properties. The attention mechanism computes a weighted sum of the context vectors:
$$\text{Attention}(Q, K, V) = \text{softmax}\left(\frac{QK^T}{\sqrt{d_k}}\right)V$$

where $d_k$ is the scaling factor. This attention output is fused with the original IP representation via a residual connection and layer normalization to produce the final, context-enriched feature vector $\mathbf{h}_{fused}$:
$$\mathbf{h}_{fused} = \text{LayerNorm}(\mathbf{z}_{ip} + \text{MultiHeadAttn}(Q, K, V))$$

This mechanism ensures that relevant semantic signals are amplified and irrelevant noise is suppressed relative to the specific IP being analyzed.

\subsubsection{Classification Head:}

\begin{figure*}[t]
    \centering
    \includegraphics[width=\textwidth]{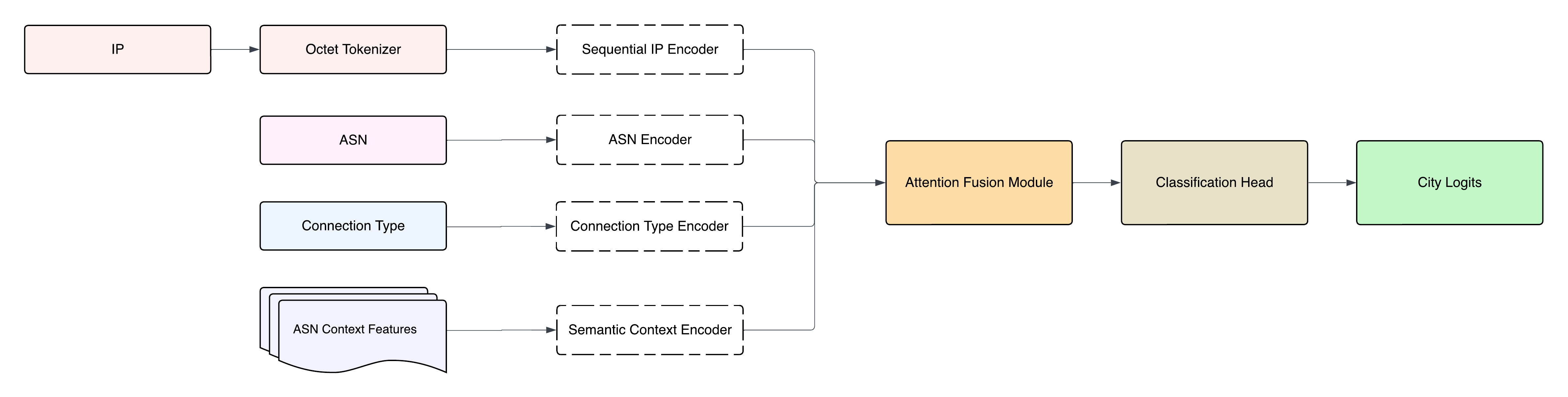}
    \caption{IPGeoAI - Architecture}
    \label{fig:overall_architecture}
\end{figure*}

The context-enriched feature vector $\mathbf{h}_{fused}$ is projected onto the output space of target City IDs via a fully connected layer $\mathbf{W}_{out} \in \mathbb{R}^{|\mathcal{C}| \times d}$:
$$\mathbf{z}_{logits} = \mathbf{W}_{out} \mathbf{h}_{fused} + \mathbf{b}_{out}$$
where $|\mathcal{C}| \approx 200,000$ represents the cardinality of unique cities. The model is optimized using Cross-Entropy Loss with label smoothing ($\epsilon=0.1$) to prevent overconfidence and improve generalization on the long tail of sparse locations. The overall architecture is shown in figure \ref{fig:overall_architecture}.

Since the country-level predictions for an IP inherently achieve higher accuracy than fine-grained city classifications, relying on the top city prediction to infer the country ('upleveling') is suboptimal. This approach discards the more robust country signal, frequently associating IPs with a wrong country. To mitigate this, we employ a constrained top-k selection strategy to be able to construct an accurate geo-hierarchy. 

Let $\mathcal{C}_{topK}$ be the set of $K=100$ cities with the highest predicted logits from the model. 
We adopt a hierarchical inference strategy. Recognizing that country-level geolocation is a mature domain with high reliability, we treat it as a coarse-grained constraint. This allows IPGeoAI to focus its capacity on the significantly harder task of resolving intra-country topological ambiguities. 
Let $c_{target}$ be the country code of the IP address as determined by a leading 3rd-party commercial vendor. The final predicted city $\hat{y}$ is selected by filtering the top candidates to match the target country:$$\mathcal{C}_{valid} = \{ c \in \mathcal{C}_{topK} \mid \text{Country}(c) = c_{target} \}$$$$\hat{y} = \arg\max_{c \in \mathcal{C}_{valid}} (\mathbf{z}_{logits})_c$$

If $\mathcal{C}_{valid}$ is empty, the system falls back to the unconstrained top prediction. This hybrid approach leverages the model's superior city-level discrimination while using the external country signal as a hard constraint to enforce geographic consistency. 

\section{Model Serving}
\label{sec:model_serving}

IPGeoAI was integrated into a high-throughput production system where real-time location queries are served using data derived from a daily batch processing pipeline. This system underpins critical personalized services, and supports downstream applications that require location data. In operation, IPGeoAI reliably processes a substantial volume of traffic, demonstrating a material improvement in the semantic precision of location signals utilized by downstream applications.

\paragraph{Inference Framework}

To ensure both scalability and prediction currency, IPGeoAI has a decoupled inference framework. While the core model operates on a static training checkpoint, inference is executed on a daily cadence, processing daily observed IP addresses and picking the latest prediction to continuously expand the cumulative prediction repository. To enhance semantic understanding, we integrate domain-specific features derived from ASN metadata; these features are refreshed weekly using an LLM, optimizing the trade-off between computational cost and feature freshness. The resulting city-level predictions are automatically aggregated into region and country-level hierarchies and indexed within the real-time user location serving infrastructure, ensuring low-latency retrieval for all downstream applications.

\section{Experiments}
\label{sec:experiments}
To rigorously assess the efficacy of IPGeoAI, we conducted a comprehensive evaluation using the ground truth datasets defined previously, spanning both offline benchmarks and live online environments across Facebook and Instagram. Our evaluation framework focuses on two primary dimensions: model recall and downstream relevance in 1st-tier metric. We begin with an offline evaluation, where IPGeoAI is trained 
% with the hyper parameters [Batch Size - 2048, Learning Rate - 0.00005, Optimizer - Adam, Number of GPUs - 64 H100s, Number of Training Epochs -     10] 
and benchmarked against Meta’s existing heuristic models, the third-party provider, and several model variants detailed in our Ablation Study. To test these findings in a production environment, we subsequently deployed IPGeoAI within our location intelligence ecosystem (specifically the "Current Location" platform) to support location-based filtering for downstream use cases.

\subsection{Baseline}
We use an external vendor as our baseline and compare IPGeoAI model predictions against the output provided by the baseline dataset here after referred to as Third Party Baseline. We also evaluated other commercial and open-source datasets but excluded them from the final benchmark due to insufficient coverage for our specific hyperscale production traffic; preliminary analysis indicated that these alternatives covered significantly fewer IPv6 ranges compared to the selected 3rd-party baseline provider. While the Machine Learning approaches described in section \ref{sec:related_work} define the current academic state-of-the-art by explicitly modeling internet topology, we deliberately excluded them from our production evaluation for three reasons -- inference scalability, dependency on active measurement, cold start problem. 

\subsubsection{Inference Latency and Scalability} State-of-the-art Graph Neural Networks (GNNs) typically require dynamic neighborhood aggregation during inference. For a $K$-layer GNN, inference complexity scales with the size of the queried node's neighborhood $O(N_{neighbors}^K)$. In a production environment serving millions of queries per second (QPS) with strict millisecond-level latency budgets (SLAs), the I/O overhead of fetching neighbor features in real-time is prohibitive. In contrast, IPGeoAI operates as a standalone sequence model with $O(1)$ complexity relative to the network graph size. This ensures deterministic, constant-time inference regardless of network connectivity, making it the only viable architecture for hyperscale deployment.

\subsubsection{Dependency on Active Measurement} High-fidelity graph construction generally relies on active probing (e.g., traceroutes, ping measurements) to establish ground-truth edges between nodes. This introduces significant operational complexity and network overhead. Our proposed architecture adheres to a strictly passive design philosophy; it requires only the target IP address and publicly available ASN metadata. This eliminates the need for maintaining a massive, constantly updating topology graph or deploying intrusive measurement infrastructure.

\subsubsection{Cold Start Scenarios} GNNs often struggle with "cold start" scenarios for nodes that lack established graph connections (e.g., new IPv6 allocations or transient mobile IPs). By treating the IP address as a hierarchical sequence enriched with semantic context, our model can generalize to these unseen or isolated subnets based on latent allocation patterns alone, without requiring neighbor information.

\subsubsection{Exclusion of Tree-Based Baselines}                                                              
We attempted to benchmark against Gradient Boosted Decision Trees (LightGBM, XGBoost), which are the industry standard for tabular data.   However, we found these methods computationally intractable for our problem space. GBDT multi-class implementations require storing and updating K gradient statistics per leaf node, causing memory to scale with O(K × leaves × trees). For $K \approx 200,000$ classes, this exceeded 256GB even with shallow trees (depth 8, 200 trees). In contrast, the Transformer computes a single d-dimensional embedding (d = 256) before  projecting to class logits, concentrating computation in the embedding layers rather than the output space. This architectural difference enables efficient handling of high-cardinality geolocation targets and scales naturally to finer granularities (zip codes, tiles). 

\subsection{IPGeoAI Results}
To evaluate our model, we focus on maximizing the total number of correct predictions. Therefore, we use Accuracy (mathematically equivalent to micro-average recall and micro-average precision in our setup with 100\% coverage) as our primary metric. Our evaluation confirms that IPGeoAI yields superior performance compared to existing baselines in both offline benchmarks and online production settings. Specifically, relative to 3rd-party provider, our model demonstrated a 6\% absolute performance improvement in City-level accuracy (from 30\% for the third party baseline to 36\% for IPGeoAI model) and a ~3\%improvement in Region-level accuracy (from 76\% for the third party baseline to 79\% for IPGeoAI model) on the designated test set.

In addition to exact IP accuracy (overall), we report City Accuracy @ 100km (percentage of test samples where the geodesic distance between the predicted city's center (centroid) and the true location is less than or equal to 100 kilometers) and stratify results by protocol (IPv4 vs. IPv6). We further evaluate model performance at broader prefix levels, specifically 'IP Trunk' (IPv4 /28, IPv6 /64) and 'IP Trim' (IPv4 /24, IPv6 /48). We evaluate these metrics at both the city level and the region level as shown in Table \ref{tab:geo_metrics}, where regional labels are derived from city predictions using the geographic hierarchy described in Section \ref{sec:geo_hierarchy_data}. To strictly evaluate topological resolution, IPGeoAI predictions were constrained by the country signal from the baseline. This ensures that the reported gains are driven by superior city-level discrimination (e.g., distinguishing San Francisco from San Jose) rather than differences in country-level classification. As shown in Table \ref{tab:geo_metrics}, 3rd-party provider retains an accuracy advantage on legacy IPv4 addresses (45.01\% vs 36.87\%). This is expected, as IPv4 is a static, saturated address space that heuristic databases have effectively memorized over decades of manual curation. However, the modern internet is increasingly defined by the dynamic, expanding IPv6 address space, where heuristic coverage is sparse. IPGeoAI significantly outperforms the baseline on IPv6 (36.46\% vs 29.58\% City Accuracy), demonstrating superior generalization on unseen subnets. Crucially, our online production tests (Section \ref{sec:online_ab_testing}) reveal that this improvement in the 'long tail' of dynamic IPs drives a +0.35\% lift in our 1st-tier downstream use cases metric which is computed by aggregating performance signals from widespread deployment across the platform . This result validates a key hypothesis: incremental benefits are no longer generated by marginally improving static IPv4 lookups, but by correctly resolving the complex, transient IPv6 and mobile networks where traditional methods fail.

\begin{table}[h]
    \centering
    \caption{IPGeoAI vs. 3rd-party provider Accuracy Comparison}
    \label{tab:geo_metrics}
    \setlength{\tabcolsep}{4pt} % Reduces space between columns slightly
    % Resizebox ensures the table fits within the column width automatically
    \resizebox{\columnwidth}{!}{%
    \begin{tabular}{lcccccc}
        \toprule
        & \multicolumn{3}{c}{\textbf{Third-Party Baseline}} & \multicolumn{3}{c}{\textbf{IPGeoAI}} \\
        \cmidrule(lr){2-4} \cmidrule(lr){5-7}
        \textbf{Metric} & \textbf{All} & \textbf{IPv4} & \textbf{IPv6} & \textbf{All} & \textbf{IPv4} & \textbf{IPv6} \\
        \midrule
        Region Accuracy (Exact) & 76.63\% & 80.00\% & 76.51\% & 79.21\% & 68.35\% & 79.58\% \\
        City Accuracy (Exact)   & 30.09\% & 45.01\% & 29.58\% & 36.47\% & 36.87\% & 36.46\% \\
        City Accuracy @ 100km    & 30.15\% & 45.14\% & 29.60\% & 36.82\% & 37.82\% & 36.78\% \\
        \bottomrule
    \end{tabular}%
    }
\end{table}

\subsection{Ablation Study}

To validate our architectural decisions, we incrementally evaluated the impact of model choice, input representation, and feature fusion strategies. Table \ref{tab:ablation_study} summarizes the progression of results.

\paragraph{Multilayer Perceptron (MLP):} As a baseline, we trained a deep MLP using IP addresses as static binary vectors concatenated with metadata (ASN and Connection Type) embeddings. This architecture achieved a test recall of 31.76\%.

\paragraph{Static Feature Transformer:} Replacing the MLP with a Transformer Encoder, while retaining the static input representation, yielded a marginal improvement to 32.19\%. This suggests that while self-attention optimizes feature interactions better than feed-forward layers, the static binary representation remains a bottleneck.

\paragraph{Sequential Octet Transformer:} Redesigning the input to treat IPs as hierarchical sequences of octets (rather than flat bit-vectors) improved recall to 33.36\%. This confirms that sequential modeling provides a superior inductive bias for capturing the hierarchical subnet allocation policies inherent to internet routing.

\paragraph{Attention-Based Feature Fusion:} Replacing static concatenation with a Multi-Head Attention mechanism (using the IP sequence as Query and metadata as Keys/Values) provided a significant gain, reaching 35.7\%. This demonstrates the necessity of dynamically weighting metadata features based on the specific network context.

\paragraph{Semantic Enrichment:} Finally, injecting LLM-derived semantic features into the fusion layer achieved our highest recall of 36.47\%. This confirms that resolving complex topological ambiguities requires both a dynamic fusion mechanism and a semantically rich feature space.

\begin{table}[h]
    \centering
    \caption{Ablation Study Results}
    \label{tab:ablation_study}
    \begin{tabular}{lc}
        \toprule
        \textbf{Model Configuration} & \textbf{Recall (\%)} \\
        \midrule
        Multi Layer Perceptron (Baseline) & 31.76 \\
        Static Feature Transformer & 32.19 \\
        Sequential Octet Transformer & 33.36 \\
        Attention-Based Feature Fusion & 35.70 \\
        \textbf{Semantic Enrichment (Ours)} & \textbf{36.47} \\
        \bottomrule
    \end{tabular}
\end{table}

\subsection{Online A/B Testing}
\label{sec:online_ab_testing}
To rigorously evaluate the effectiveness of our IPGeoAI model on live traffic, we conducted large-scale online A/B test on our platforms. The experiment was designed to measure the real-world impact of the model over a sustained period. The test ran for 13 consecutive days where users were randomly assigned to either the control group (serving the legacy waterfall model) or the treatment group (serving the waterfall with IPGeoAI model). The quality of location intelligence directly correlates with the efficiency of our  ecosystem. Evaluated across a diverse ecosystem of downstream applications, the model achieved a +0.35\% lift in our 1st-tier downstream use cases metric. This statistically significant improvement validates the model's robustness and its ability to drive value across heterogeneous production environments.

\section{Conclusion and Future Work}
\label{sec:conclusion}
We presented IPGeoAI, a novel deep learning framework based on the Transformer Architecture for high-precision IP Geolocation addressing the critical industry challenge of resolving network addresses to city-level locations at scale. We demonstrated that the latent hierarchical structure of IP addresses can be effectively captured by a Transformer architecture. Crucially, we showed that the integration of unstructured semantic context—derived from LLMs—is the key to resolving the geographic ambiguities that confound traditional systems. Our rigorous ablation study confirmed that while sequential modeling provides a strong numerical foundation, it is the Multi-Head Attention Fusion of semantic metadata that unlocks state-of-the-art performance. This architecture not only outperforms heuristics baselines and third party vendors on real world traffic but provides a scalable, generalizable solution for new and dynamic addresses where historical data is absent. The enhanced performance directly translated to an improvement in our 1st-tier downstream use cases metric demonstrating the viability of our approach in a hyperscale production environment.

Our current findings establish IPGeoAI as a robust framework for IP-based geolocation; however, several avenues remain for future exploration to address the evolving demands of hyperscale location intelligence. While our current batch-inference architecture serves daily updates efficiently, the next frontier is enabling purely real-time inference. We plan to optimize the model for low-latency serving to handle peak loads traffic. Currently, our model optimizes for city-level precision and fetches country from a third party dataset to not degrade country accuracy. We also aim to build models that serve the entire geo-hierarchy from country to zip and even hyper local boundary tiles.

%%%%%%%%%%%%%%%%%%%%%%%%%%%%%%%%%%%%%%%%%%%%%%%%%%%%%%%%%%%%%%%%%%%%%%%%%%%%%%%%%%%%%%%%%%%%%%%
\bibliographystyle{ACM-Reference-Format}
\bibliography{references}
%%%%%%%%%%%%%%%%%%%%%%%%%%%%%%%%%%%%%%%%%%%%%%%%%%%%%%%%%%%%%%%%%%%%%%%%%%%%%%%%%%%%%%%%%%%%%%%

\appendix

\section{Hyperparameters and Training Details}

\label{app:hyperparams}
We provide the complete specification of the model architecture and optimization hyperparameters (Table \ref{tab:infra_config} and Table \ref{tab:optim_config}).

\begin{table}[h]
    \centering
    \caption{Infrastructure \& Model Configuration}
    \begin{tabular}{ll}
        \toprule
        \textbf{Parameter} & \textbf{Value} \\
        \midrule
        \multicolumn{2}{l}{\textit{Infrastructure}} \\
        Compute Nodes & 8 $\times$ (8 NVIDIA H100 GPUs) \\
        Parallel Strategy & Fully Sharded Data Parallel (FSDP) \\
        Precision & Mixed Precision (BF16/FP32) \\
        \midrule
        \multicolumn{2}{l}{\textit{Architecture (Transformer Encoder)}} \\
        Layers ($N$) & 16 \\
        Attention Heads & 8 \\
        Embedding Dim ($d_{model}$) & 256 \\
        Feedforward Dim ($d_{ff}$) & 1024 \\
        Dropout & 0.2 \\
        Output Classes & 200,000 (City IDs) \\
        \bottomrule
    \end{tabular}
    \label{tab:infra_config}
\end{table}

\begin{table}[h]
    \centering
    \caption{Optimization Parameters}
    \begin{tabular}{ll}
        \toprule
        \textbf{Hyperparameter} & \textbf{Value} \\
        \midrule
        Optimizer & AdamW \\
        Peak Learning Rate & $5.0 \times 10^{-5}$ \\
        Weight Decay & $1.0 \times 10^{-4}$ \\
        Batch Size & 2048 \\
        Total Epochs & 10 \\
        Scheduler & OneCycleLR (Cosine Annealing) \\
        Warmup Steps & 5\% of total steps \\
        \bottomrule
    \end{tabular}
    \label{tab:optim_config}
\end{table}

\section{Qualitative Analysis of LLM Predictions}

We conducted a manual audit of 25 randomly sampled ASNs to validate the "soft labels" generated by the LLM. Table \ref{tab:llm_audit} presents a subset of these samples, comparing the LLM prediction against ground truth derived from PeeringDB and manual investigation. The audit revealed a 80\% agreement rate, with errors primarily occurring in ambiguous "Mixed" / "Unknown" usage scenarios.

\begin{table*}[h]
    \centering
    \caption{LLM Prediction Audit Examples}
    \begin{tabular}{p{0.08\linewidth} p{0.2\linewidth} p{0.25\linewidth} p{0.2\linewidth} c}
        \toprule
        \textbf{ASN ID} & \textbf{ASN Name} & \textbf{LLM Prediction} & \textbf{Ground Truth} & \textbf{Valid?} \\
        \midrule
        \textit{27198} & \textit{Indiana University} & \texttt{Sector: Education} \newline \texttt{Type: University} & Education / University & \checkmark \\
        \midrule
        \textit{13848} & \textit{Ariba Inc.} & \texttt{Sector: Private Enterprise} \newline \texttt{Type: Enterprise} & Enterprise / Private & \checkmark \\
        \midrule
        \textit{13534} & \textit{Ralls Technologies, LLC} & \texttt{Sector: Private Enterprise} \newline \texttt{Type: Other} & Enterprise / Private & $\times$ \\
        \midrule
        \textit{27145} & \textit{DoD Network Information Center} & \texttt{Sector: Government} \newline \texttt{Type: Government} & Government / Government & \checkmark \\
        \bottomrule
    \end{tabular}
    \label{tab:llm_audit}
\end{table*}

\section{Inference Logic Walkthrough}

Based on the methodology described in Section \ref{sec:model_architecture}, here is a concrete walkthrough of the constrained inference logic. This example demonstrates how the model predicts a CityID for a given IP address. City names are used here for simplicity.

\subsection*{Scenario}
The model receives an input IP address (\texttt{2001:db8::1234}) and must predict the correct City ID. The external country signal indicates the IP belongs to the United States. This setup reflects our design choice to use external signals for macro-level filtering while relying on IPGeoAI for granular city-level resolution.

\begin{itemize}
    \item Input Data:
    \begin{itemize}
        \item IP Address: \texttt{2001:db8::1234}
        \item External Signal (3rd-party provider): Country = \texttt{US} (United States)
    \end{itemize}

    \item Raw Model Output (Top-K Selection):
    The Classification Head outputs logits for all 200,000+ cities. We select the top $K=3$ (simplified for this example; actual $K=100$) based on raw probability as shown in table \ref{tab:inference_example}.

    \begin{table}[h]
        \centering
        \caption{Raw Logit Probabilities (Before Constraint)}
        \begin{tabular}{lllc}
            \toprule
            \textbf{Rank} & \textbf{City Name} & \textbf{Country} & \textbf{Raw Probability} \\
            \midrule
            1 & Paris & France (\texttt{FR}) & \textbf{0.65} \\
            2 & Lyon & France (\texttt{FR}) & 0.15 \\
            3 & Paris & United States (\texttt{US}) & \textbf{0.12} \\
            \bottomrule
        \end{tabular}
        \label{tab:inference_example}
    \end{table}

    \textit{Note: Without the constraint logic, the model would incorrectly predict ``Paris, France'' because it has the highest raw probability.}

    \item Constraint Application:
    We construct the valid candidate set $C_{valid}$ by filtering the Top-K list against the target country $c_{target} = \texttt{US}$.

    \begin{itemize}
        \item Candidate 1: Paris, FR $\rightarrow$ \textsc{Reject} (Country \texttt{FR} $\neq$ \texttt{US})
        \item Candidate 2: Lyon, FR $\rightarrow$ \textsc{Reject} (Country \texttt{FR} $\neq$ \texttt{US})
        \item Candidate 3: Paris, US $\rightarrow$ \textsc{Keep} (Country \texttt{US} $==$ \texttt{US})
    \end{itemize}

    \item Final Selection:
    The system selects the candidate with the highest probability from the filtered $C_{valid}$ set.

    \begin{itemize}
        \item Final Prediction: Paris, United States (Probability: 0.12)
    \end{itemize}

    \item Fallback Mechanism:
    In the rare event that $C_{valid}$ is empty (i.e., no cities in the Top-100 match the target country), the system falls back to the unconstrained top prediction (Paris, France) to ensure a prediction is always returned.
\end{itemize}

\end{document}